\documentclass{article}

\usepackage{arxiv}

\usepackage[utf8]{inputenc} 
\usepackage[T1]{fontenc}    
\usepackage{hyperref}       
\usepackage{url}            
\usepackage{booktabs}       
\usepackage{amsfonts}       
\usepackage{nicefrac}       
\usepackage{microtype}      
\usepackage{lipsum}		
\usepackage{amsthm}
\usepackage{graphicx}
\usepackage{doi}
\usepackage{physics}
\usepackage{amssymb}
\usepackage{pifont}%

\newcommand{\pd}[2]{\frac{\partial #1}{\partial #2}}

\usepackage{setspace}
\title{Finding geodesics with the Deep Ritz method}

\date{} 					

\author{
    Conor Rowan \\
	Smead Aerospace Engineering Sciences\\
	University of Colorado Boulder\\
    3775 Discovery Drive\\
	Boulder, CO 80309 \\
	\texttt{conor.rowan@colorado.edu} \\
}

\renewcommand{\headeright}{}
\renewcommand{\undertitle}{}
\renewcommand{\shorttitle}{}

\hypersetup{
pdftitle={A template for the arxiv style},
pdfsubject={q-bio.NC, q-bio.QM},
pdfauthor={David S.~Hippocampus, Elias D.~Striatum},
pdfkeywords={First keyword, Second keyword, More},
}

\begin{document}
\maketitle

\begin{abstract}
    Geodesic problems involve computing trajectories between prescribed initial and final states to minimize a user-defined measure of distance, cost, or energy. They arise throughout physics and engineering---for instance, in determining optimal paths through complex environments, modeling light propagation in refractive media, and the study of spacetime trajectories in control theory and general relativity. Despite their ubiquity, the scientific machine learning (SciML) community has given relatively little attention to investigating its methods in the context of these problems. In this work, we argue that given their simple geometry, variational structure, and natural nonlinearity, geodesic problems are particularly well-suited for the Deep Ritz method. We substantiate this claim with four numerical examples drawn from path planning, optics, solid mechanics, and generative modeling. Our goal is not to provide an exhaustive study of geodesic problems, but rather to identify a promising application of the Deep Ritz method and a fruitful direction for future SciML research.
\end{abstract}




\section{Introduction}

\paragraph{} A geodesic is a curve connecting two states which minimizes a problem-specific notion of distance, cost, or energy. An elegant pedagogical introduction to geodesics is the brachistochrone, which is a curve of minimum travel time for an object sliding frictionlessly in a uniform gravitational field \cite{goldstein_classical_2002}. In path planning applications, the geodesic curve represents a trajectory of minimum distance, ensuring efficient operations for robots navigating complex environments \cite{bao_geodesic-based_2026}. Noting that the speed of light varies with the refractive index of the medium, optical geodesics are paths of minimum travel time rather than distance  \cite{russo_geodesic_1983}. In modern control theory, geodesics have helped connect ideas from differential geometry with those of optimal control \cite{sussmann_geometry_1999}. Trajectories in general relativity are computed by solving geodesic equations within spacetime geometries curved by massive bodies \cite{cranganore_einstein_2025}. Geodesic paths have also been used to smoothly deform two images into one another \cite{wu_learning_2024}. Because the problem of finding a minimal trajectory is general, it is not surprising that applications of geodesics are found across many scientific disciplines. 

\paragraph{} In many problems of practical interest, it is not possible to compute the geodesic path analytically. Numerical methods are thus required to approximate a minimum to the objective functional. A common strategy is to solve the Euler-Lagrange equations associated with the minimization problem. However, given that initial and final states of the path are prescribed, finding a geodesic represents a boundary value problem (BVP) rather than an initial value problem (IVP). This prohibits the direct application of standard time integration techniques, which require two initial conditions. To remedy this, a "shooting method" is often employed, whereby the initial velocity is iterated until the trajectory ends at the desired final state, thus solving the BVP \cite{sutti_shooting_2023, cui_continuation_2022, cho_shape_2009, noakes_finding_2022}. We remark that shooting methods incur significant computational cost by repeatedly time integrating the Euler-Lagrange equations at different settings of the initial velocity. Additional time integration is required to compute gradients of the final condition, which is a necessary component of a Newton scheme to find the desired initial velocity. In this work, we opt to formulate the geodesic problem as a static BVP. This choice allows us to treat the initial and final states as Dirichlet boundary conditions on the geodesic path, thus avoiding iteration of the initial conditions. Furthermore, we propose solving for the geodesic by directly minimizing the objective functional, rather than solving the corresponding Euler-Lagrange equations. Of course, the variational structure of geodesic problems is well-understood, but treatments which emphasize this tend to be more theoretical than numerical \cite{struwe_direct_2008, postnikov_variational_2019}. Two exceptions are \cite{schmidt_shape_2006} and \cite{meng_variational_2021}, where the shooting method is abandoned in favor of minimizing the objective with gradient descent, though, to the best of our knowledge, examples of this sort are rare. We note for future reference that when the trajectory is discretized in a basis and computed directly through, we obtain a so-called "Ritz" method. Additionally, given that the geodesic is governed by a minimization principle, we say that it is "variational." 

\paragraph{} Despite their resemblance to variational BVPs in engineering mechanics, the scientific machine learning (SciML) community has not explored its methods in the context of geodesic problems. The discipline of SciML got its start in the period of 2017-2019 with the introduction of physics-informed neural networks (PINNs) and the Deep Ritz method (DRM) \cite{sirignano_dgm_2018, raissi_physics-informed_2019, e_deep_2017}. These methods discretize the solution of an ordinary or partial differential equation with a neural network and determine the parameters of the network by incorporating the governing equations into the loss function. In the case of PINNs, the integral of the pointwise squared error of the strong form of the differential equation is used as an objective. In contrast, DRM exploits variational structure by minimizing a discretized energy functional in order to solve the corresponding Euler-Lagrange equations. These methods have been studied in many application areas such as contact mechanics \cite{sahin_solving_2024}, heat transfer \cite{madir_physics_2024}, fluid mechanics \cite{jin_nsfnets_2021}, hyperelasticity \cite{abueidda_deep_2022}, and fracture mechanics \cite{manav_phase-field_2024}. To the best of the author's knowledge, there are no existing works using PINNs or DRM to find geodesics.

\paragraph{} One takeaway from the PINNs and DRM literature is that neural network discretizations of BVPs are most effective on problems that satisfy three criteria \cite{rowan_variational_2025}. The first criterion is that the problem geometry is simple, so that boundary conditions can be enforced explicitly without a mesh. The second is that there is a natural way to phrase the solution as a minimization problem. The third is that the problem is inherently nonlinear, such that the neural network discretization is not the only reason iterative solution methods are required. Given the one-dimensional time-like geometry of the geodesic paths, their inherently variational structure, and the nonlinearity introduced by the dependence of the metric on position, we posit that geodesic problems satisfy these criteria and are thus good candidates for neural network discretizations. This work represents an initial exploration of geodesic problems for the SciML community. Treating it as a static BVP, we discretize the geodesic problem with a neural network in such a way that the boundary conditions are enforced automatically, allowing us to use DRM without constraints. We demonstrate the ease and efficacy of this strategy on four example problems. The rest of this work is organized as follows. In Section 2, we outline the variational and strong forms of the geodesic boundary value problem for motion on an arbitrary parameterized surface. In Section 3, we discuss the Deep Ritz formulation and a method to enforce Dirichlet boundary conditions automatically. In Section 4, we provide four numerical examples from path planning, optics, solid mechanics, and generative modeling to illustrate our methodology. In Section 5, we close with concluding remarks and directions for future work.


\section{Variational geodesic equation}

\paragraph{} Consider a parameterized surface $\mathbf{N}(\boldsymbol \theta): \mathbb{R}^k \rightarrow \mathbb{R}^n$ where $k$ defines the intrinsic dimension of the surface and $n$ is the dimension of the ambient space in which it is embedded. Suppose that the parameters $\boldsymbol \theta$ which traverse the surface are themselves parameterized with a time-like variable $t \in [0,T]$, thus defining a path in parameter space $\boldsymbol \theta(t)$ and a path in "physical" space $\mathbf{N}(\boldsymbol \theta(t))$. For the given parameterization $\boldsymbol \theta(t)$, we wish to compute the distance traveled along this path in physical space. We choose an energy-like distance measure, which is computed as the time integral of the squared magnitude of the instantaneous velocity \cite{hartle_gravity_2021}. This is given by

\begin{equation}\label{length}
    E\Big( \boldsymbol \theta(t) \Big ) = \frac{1}{2} \int_0^T \frac{d N_i}{dt} \cdot \frac{d N_i}{dt} dt = \frac{1}{2}\int_0^T \pd{N_i}{\theta_j} \pd{\theta_j}{t} \pd{N_i}{\theta_k} \pd{\theta_k}{t} dt = \frac{1}{2} \int_0^T g_{jk}(\boldsymbol \theta) \pd{\theta_j}{t} \pd{\theta_k}{t} dt.
\end{equation}

The metric tensor $\mathbf{g}$ encodes the geometry of the problem and is defined as 

\begin{equation}\label{metric}
    g_{jk}(\boldsymbol \theta) = \pd{\mathbf{N}}{\theta_j} \cdot \pd{\mathbf{N}}{\theta_k}.
\end{equation} 

Note that Eq. \eqref{length} differs from the arc length of the path by a square root in the integrand. We remark that the true arc length is independent of the speed at which the path is traversed, whereas the value of the "energy" in Eq. \eqref{length} depends not just on the path but also on the rate of travel along the path. This point is immaterial, as the minimum energy and minimum arc-length paths coincide, even if they may not agree on the velocities. Thus, with a clean conscience, the energy measure is adopted throughout to ease calculations.

\paragraph{} When the parameterized path $\boldsymbol \theta(t)$ is given, Eq. \eqref{length} provides a recipe to compute a measure of distance along the path. A more interesting problem is to take only the endpoints as known and to find the shortest path on the surface between them. In this case, the initial and final parameter settings $\boldsymbol \theta(0)$ and $\boldsymbol \theta(T)$ are given and the rest of $\boldsymbol \theta(t)$ must be determined. To solve this problem, Eq. \eqref{length} is converted into a variational boundary value problem:

\begin{equation}\label{varbvp}
\begin{aligned}
    \underset{\boldsymbol \theta(t)}{ \text{argmin }} E\Big( \boldsymbol \theta(t) \Big ) \\
    \text{s.t. } \boldsymbol \theta(0) = \boldsymbol \theta_0, \quad \boldsymbol \theta(T) = \boldsymbol \theta_T.
\end{aligned}
\end{equation}

A solution to Eq. \eqref{varbvp} corresponds to a geodesic on the parameterized surface $\mathbf{N}(\boldsymbol \theta)$. A Ritz method works with the minimization problem directly, but a common strategy in the literature is to solve the corresponding Euler-Lagrange equations. Taking a variation of Eq. \eqref{length}, we obtain

\begin{equation*}
    \delta E = \int_0^T \frac{1}{2} \pd{g_{jk}}{\theta_{\ell}} \dot \theta_j \dot \theta_k \delta \theta_{\ell} + g_{j\ell} \dot \theta_j \delta \dot \theta_{\ell} dt = \int_0^T \frac{1}{2} \pd{g_{jk}}{\theta_{\ell}} \dot \theta_j \dot \theta_k \delta \theta_{\ell} - \pd{}{t}(g_{j\ell} \dot \theta_j )\delta \theta_{\ell} dt = 0.
\end{equation*}

In the second equality, we have performed integration by parts and used that $\delta \theta(0) = \delta \theta(T) = 0$. Because the variation $\delta \boldsymbol \theta$ is arbitrary, the integral is only zero when the coefficient of the variation is zero pointwise. Such a requirement yields the Euler-Lagrange equations for the geodesic:

\begin{equation*}
    \frac{1}{2} \pd{g_{jk}}{\theta_{\ell}} \dot \theta_j \dot \theta_k - \pd{g_{j\ell}}{\theta_k} \dot \theta_k \dot \theta_j - g_{\ell j} \ddot \theta_j = 0.
\end{equation*}

Note that the metric tensor is symmetric, which allows us to write $g_{j\ell} = g_{\ell j}$. Multiplying through by the inverse metric tensor $g^{-1}_{i\ell}$, we obtain

\begin{equation*}
    \ddot \theta_i + g^{-1}_{i \ell}\qty( \pd{g_{j \ell}}{\theta_k} - \frac{1}{2} \pd{g_{jk}}{\theta_{\ell}}) \dot \theta_k \dot \theta_j = 0.
\end{equation*}

It is common at this point to define the Christoffel symbols in order to simplify this equation. The Christoffel symbols are given by 

\begin{equation*}
    \Gamma_{ijk} = g^{-1}_{i \ell}\qty( \pd{g_{j \ell}}{\theta_k} - \frac{1}{2} \pd{g_{jk}}{\theta_{\ell}}).
\end{equation*}

Note that other texts may use symmetries of the above quantities to define the Christoffel symbols differently \cite{hartle_gravity_2021}, but this expression suffices for the purposes of exposition. With this definition, the geodesic equation becomes 

\begin{equation}\label{geo2}
    \ddot \theta_i + \Gamma_{ijk} \dot \theta_j \dot \theta_k = 0, \quad \boldsymbol \theta(0)=\boldsymbol \theta_0, \quad \boldsymbol \theta(T) = \boldsymbol \theta_T.
\end{equation}

We note that this is a boundary value problem despite the time-like curve parameter $t$. As such, it is not possible to solve Eq. \eqref{geo2} with a standard time-stepping algorithm that marches forward in time. As noted above, many implementations use shooting methods, which iteratively adjust the initial velocity until the final condition is met. In this work, we treat Eqs. \eqref{varbvp} and \eqref{geo2} as BVPs, thus avoiding expensive iterations of the shooting method.


\section{Deep Ritz method for geodesics}

\paragraph{} Using the Deep Ritz method \cite{e_deep_2017}, we discretize the path in parameter space $\boldsymbol \theta(t)$ with a neural network and determine the geodesic with the minimization problem of Eq. \eqref{varbvp}. Without loss of generality, we assume that the time elapsed along the path is $T=1$. The path is discretized as 

\begin{equation}\label{distance}
    \hat{\boldsymbol{\theta}}(t; \boldsymbol \beta) = \boldsymbol \theta_0(1-t) + \boldsymbol \theta_1 t + \sin(\pi t) \mathcal{N}(t;\boldsymbol \beta),
\end{equation}

\noindent where $\mathcal{N}: \mathbb{R} \rightarrow \mathbb{R}^k$ is a neural network with trainable parameters $\boldsymbol \beta$ that need not satisfy the initial and final conditions \cite{wang_exact_2023, sukumar_exact_2022, sheng_pfnn_2021}. By treating the parameter $t$ analogously to the spatial coordinate in a static boundary value problem, this "distance function" approach enforces the boundary conditions automatically. The Deep Ritz formulation of the geodesic problem is then 

\begin{equation}\label{drm}
\begin{aligned}
    \hat E( \boldsymbol \beta) = \frac{1}{2} \int_0^1 \pd{\hat{\boldsymbol{\theta}}(t; \boldsymbol \beta)}{t} \cdot \mathbf{g}\Big( \hat{\boldsymbol{\theta}}(t; \boldsymbol \beta) \Big) \pd{\hat{\boldsymbol{\theta}}(t; \boldsymbol \beta)}{t} dt\\
    \underset{\boldsymbol \beta}{\text{argmin }} \hat E(\boldsymbol \beta),
\end{aligned}
\end{equation}

\noindent where the discrete optimization problem is without constraints. We note that, given its variational structure and inherent nonlinearity, the geodesic problem represents a natural application of the Deep Ritz method. By nonlinearity, we mean that stationarity of the objective $\nabla_{\boldsymbol \beta} \hat E = \mathbf{0}$ leads to a nonlinear system of equations even for standard finite element or spectral discretizations. The only exception to this is when the metric is constant, i.e. $\mathbf{g}(\boldsymbol \theta) = \mathbf{g}_0$. Thus, in general, iterative solution methods are required for the nonlinear system. In other words, the neural network discretization does not transition a problem governed by a linear solve to a nonlinear solve. This fact is a boon if DRM is to be competitive with incumbent strategies such as finite element and spectral methods.

\paragraph{} In all numerical examples, we use ADAM to solve the optimization problem of Eq. \eqref{drm} with a learning rate of $5 \times 10^{-3}$. We employ a uniform grid of $250$ points to discretize the interval $t\in[0,1]$ and to numerically integrate the energy objective. The metric tensor is computed using symbolic or automatic differentiation. Unless otherwise indicated, we take the architecture of the network discretizing the parameter trajectory in Eq. \eqref{distance} to be a multilayer perceptron (MLP). In an MLP, the input-output relation for the $i$-th hidden layer is

\begin{equation*}
    \mathbf{y}_i = \sigma\Big(  \mathbf{W}_i\mathbf{y}_{i-1} + \mathbf{B}_i  \Big),
\end{equation*}

\noindent where $\sigma(\cdot)$ is a nonlinear activation function applied element-wise. As shown, the output $\mathbf{y}_i$ then becomes the next layer's input. The parameters of the neural network are the collection of the weight matrices $\mathbf{W}_i$ and bias vectors $\mathbf{B}_i$ for each layer. Thus, we can write the neural network parameters as $\boldsymbol{\beta} = [ \mathbf{W}_1, \mathbf{B}_1, \mathbf{W}_2,\mathbf{B}_2,\dots]$. We use a linear mapping with no bias to go from the last hidden layer to the output and hyperbolic tangent activation functions. The depth of the network and the size of the hidden layers are specified on a case-by-case basis.


\section{Numerical examples}

\subsection{Traversing a mountainous landscape}

\paragraph{} Our first numerical example illustrates a practical path planning geodesic problem which is convenient to visualize. We search for the shortest path in $\mathbb{R}^3$ through a landscape with mountains. We take the landscape to be parameterized by latitude- and longitude-type coordinates $\theta_1$ and $\theta_2$, which represent the position on a topographic map. The surface is parameterized with

\begin{equation*}
    \mathbf{N}(\boldsymbol \theta) = \begin{bmatrix} \theta_1 \\ \theta_2 \\ h \Big( \sin( f \pi \theta_1) \sin(f \pi \theta_1 \theta_2 ) \Big)^2
    \end{bmatrix},
\end{equation*}

\begin{figure}[hbt!]
\centering
\includegraphics[width=1.0\textwidth]{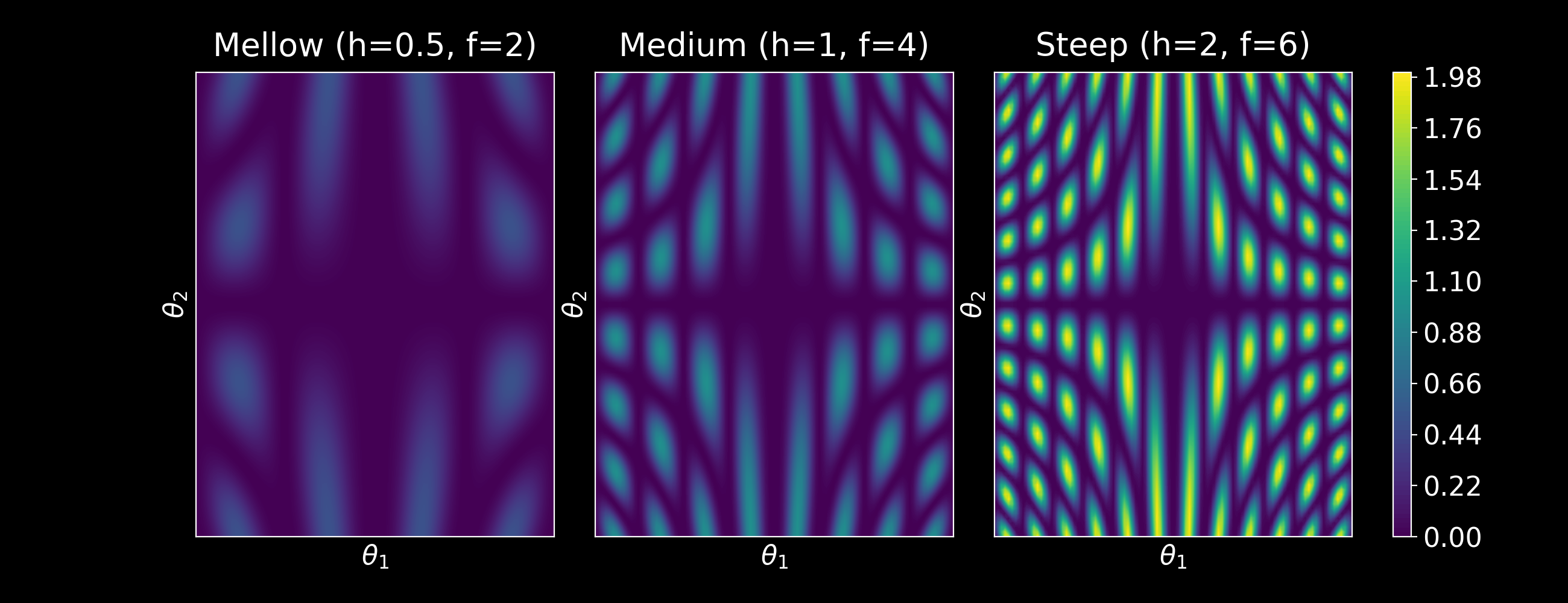}
\caption{Different types of terrain are obtained by varying the height and frequency of the hills with $h$ and $f$ respectively. Intuitively, taller and more frequent hills necessitate circuitous paths.}
\label{landscapes}
\end{figure}
\begin{figure}[hbt!]
\centering
\includegraphics[width=1.0\textwidth]{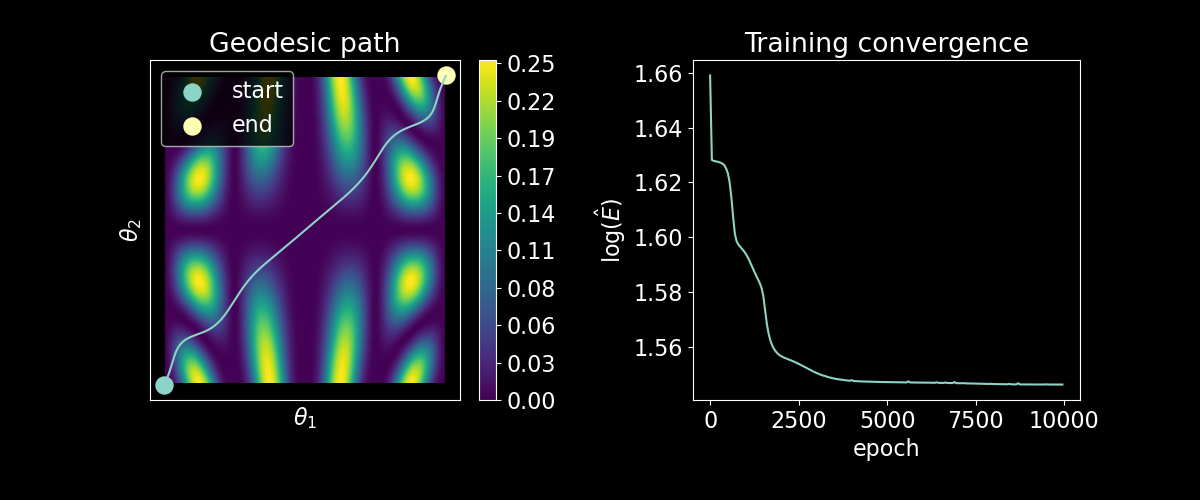}
\caption{When the landscape is mellow, approximately straight line paths are minimal distance, despite crossing over some foothills.}
\label{1ex1}
\end{figure}
\begin{figure}[hbt!]
\centering
\includegraphics[width=1.0\textwidth]{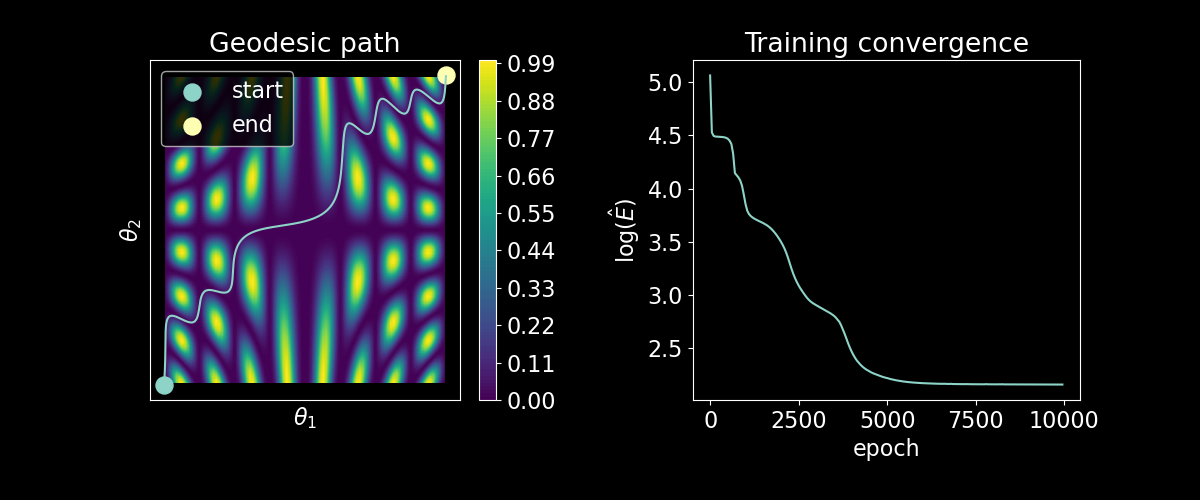}
\caption{More severe terrain requires staying in valleys, as the cost of climbing uphill to move in a straight line overshadows the cost of wandering.}
\label{1ex2}
\end{figure}


\noindent where the third component specifies the elevation and $h$ and $f$ control the severity of the terrain. See Figure \ref{landscapes} for three example landscapes generated from this parameterization. We take $\theta_1, \theta_2 \in [-1,1]$, and wish to diagonally traverse the landscape from $\boldsymbol \theta_0 = [-1,-1]^T$ to $\boldsymbol \theta_1 = [1,1]^T$ with a path of minimal length. To discretize the path, we use a two hidden layer MLP network with a width of $25$ neurons per layer. The boundary conditions are built into the discretization per Eq. \eqref{distance}, the metric is computed with Eq. \eqref{metric}, and we solve the geodesic problem with Eq. \eqref{drm} using ADAM over $1 \times 10^4$ epochs. To simulate a mellow landscape, we take $h=0.25$ and $f=2$. See Figure \ref{1ex1} for the landscape and path of minimal length. The path bends around the extreme high points of the landscape but passes over some foothills, owing to the small vertical climb required. The geodesic path in a more severe landscape given by $h=2$ and $f=4$ is shown in Figure \ref{1ex2}. Here, the shortest path is more circuitous as a result of staying in valleys to avoid steep terrain.

\subsection{Helical wave guide}

\paragraph{} A common variational problem in optics arises from Fermat's principle, which states that light rays in a medium follow the path of minimum travel time \cite{goldstein_classical_2002}. If $c$ is the speed of light in a vacuum, the refractive index of the medium $n(\mathbf{x})$ is given by

\begin{equation*}
    n(\mathbf{x}) = c / v(\mathbf{x}),
\end{equation*}

\noindent where $v(\mathbf{x})$ is the speed of light in the medium at the corresponding point in space. We parameterize the path through the medium as $\mathbf{x}(t) = \boldsymbol \theta(t)$. For a material described by a given refractive index field $n(\mathbf{x})$, the geodesic problem is

\begin{equation}\label{fermat}
    E\Big( \boldsymbol \theta(t) \Big) = \frac{1}{2}\int_0^1 n^2(\boldsymbol \theta(t) )\lVert\dot{\boldsymbol \theta} \rVert^2 dt,
\end{equation}

\noindent where the metric tensor is $\mathbf{g}(\boldsymbol \theta) = n^2(\boldsymbol{\theta}) \mathbf{I}$, indicating flat space. 

\paragraph{} As before, we discretize the path using an MLP neural network with the boundary conditions built in per Eq. \eqref{distance}. We take the computational domain to be the cube $\Omega=[-1,1]^3$ and we seek a path of minimum travel time from the center of the top surface to the center of the bottom surface. Accordingly, the initial and final conditions are given by $\boldsymbol \theta_0 = [0,0,1]^T$ and $\boldsymbol \theta_1 = [0,0,-1]^T$. The refractive index field will be that of a wave guide in which a helical tube with fast wave speeds connects the entry and exit points of the light ray. The axis of the helix can be parameterized by the height $\theta_3$:

\begin{equation}\label{helix}
    \mathbf{h}(\theta_3) = \begin{bmatrix}
        R \cos(\pi \theta_3/2)\cos( \pi \theta_3)  \\
        R \cos(\pi \theta_3/2)\sin( \pi \theta_3)  \\
        \theta_3
    \end{bmatrix},
\end{equation}

\begin{figure}[hbt!]
\centering
\includegraphics[width=1.0\textwidth]{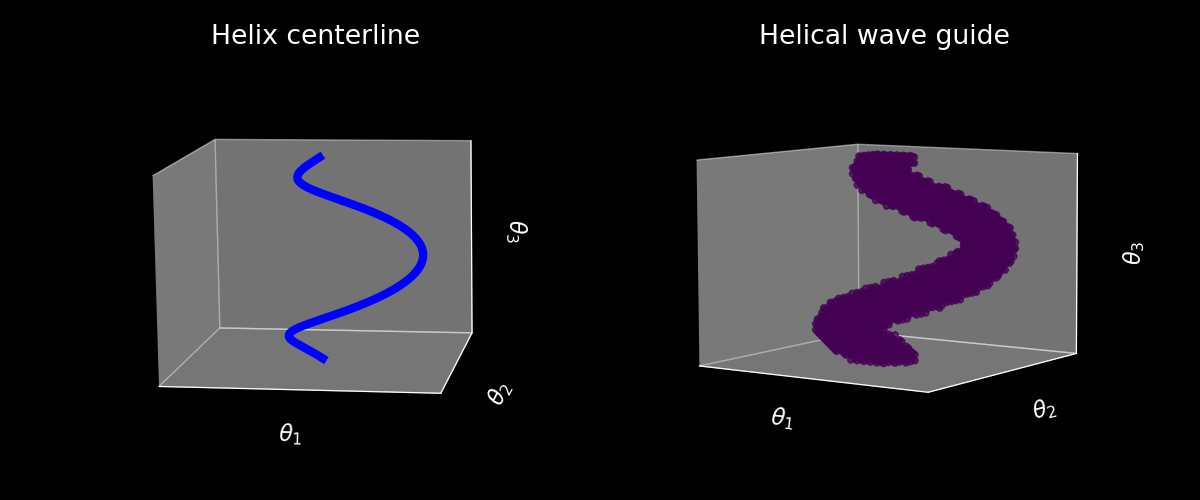}
\caption{Visualizing the centerline of the helix-shaped wave guide (left) and the helical tube in which the wave speed is large (right). Outside of the tube, the refractive index is large, indicating longer travel times.}
\label{helix_plot}
\end{figure}

\noindent where the function $\cos( \pi \theta_3/2)$ contracts the radius of the helix down to zero at the end states. Using this parameterization, the refractive index field is constructed with

\begin{equation}\label{nfield}
    n( \mathbf{x}) = \begin{cases}
        n_0, \quad  \text{if}  \underset{t}{\text{ min }} \lVert \mathbf{x} - \mathbf{h}(t)\rVert \leq \epsilon,\\
        n_1, \quad \text{else}, 
    \end{cases}
\end{equation}

\noindent where $n_0 < n_1$. Eq. \eqref{nfield} simply states that the refractive index is small in a neighborhood of radius $\epsilon$ around the helix. See Figure \ref{helix_plot} to visualize the axis of the helix and the geometry of the wave guide defined by Eq. \eqref{nfield}. In our implementation, we store the refractive index field at discrete points on a uniform grid, then train a neural network to represent this field. This provides an interpolation of the refractive index field which can be differentiated with respect to its inputs, as is required when minimizing Eq. \eqref{fermat}. Going forward, we set the radius of the helix to $R=0.75$ and the width of the wave guide to $\epsilon=0.2$.

\paragraph{} We note that the wave guide Deep Ritz problem corresponding to Eqs. \eqref{drm} and \eqref{fermat} is highly non-convex. When $n_1 \gg n_0$, the path of shortest travel time will follow the helical wave guide. However, we observe numerically that the neural network discretization of the path converges to an approximately straight line path between the initial and final states. In order to escape this local minimum, the path needs to begin winding in a helical fashion. But, because the path shape depends continuously on the neural network parameters, the winding process must happen gradually. Unfortunately, transitioning from a straight line connection to the beginning of a wound path necessarily increases the travel time, as a diagonal trajectory is taken through the slow medium. In other words, once in this local minimum, the optimizer must first significantly increase the objective value before it can find a better solution which follows the helix. In our numerical experiments, such an escape from the straight line local minimum does not occur for standard MLP networks. The problem of local minima for geodesics is noted in \cite{mortari_using_2022}. We thus explore modifications to the network architecture that make the helical solution more accessible in the loss landscape. One candidate modification is the SIREN network, which replaces standard activation functions with sinusoids \cite{sitzmann_implicit_2020}. With a SIREN, the activation at each layer is $\sigma(\cdot) = \sin(\omega_0 (\cdot))$, which naturally introduces oscillatory behavior into the discretization. Though such oscillatory behavior in some sense mirrors the parameterization of the helix, we observe that the SIREN network also converges to and stagnates at the straight line solution.

\paragraph{} Another modification of the network architecture is Fourier feature embedding, which introduces periodic behavior into the input layer of an MLP network \cite{wang_eigenvector_2021}. Whereas the standard MLP in Eq. \eqref{distance} takes in the time-like coordinate $t$, the Fourier feature network's input layer is 

\begin{equation*}
    \boldsymbol \gamma(t) = [ \sin( 2\pi \mathbf{B} t) , \cos(2\pi \mathbf{B} t)]^T,
\end{equation*}

\noindent where $\mathbf{B} \in \mathbb{R}^{f}$ is a vector whose components are normally distributed with variance $\sigma^2$ and $2f$ is the number of Fourier features. The variance of the components of the random embedding vector $\mathbf{B}$ determines the frequency content of the discretization. A standard MLP then operates on the Fourier features at the input layer. We observe from numerical experimentation that the embedded Fourier features improve the performance on the wave guide problem. This is likely due to initial path geometries which are already wound.

\begin{figure}[hbt!]
\centering
\includegraphics[width=1.0\textwidth]{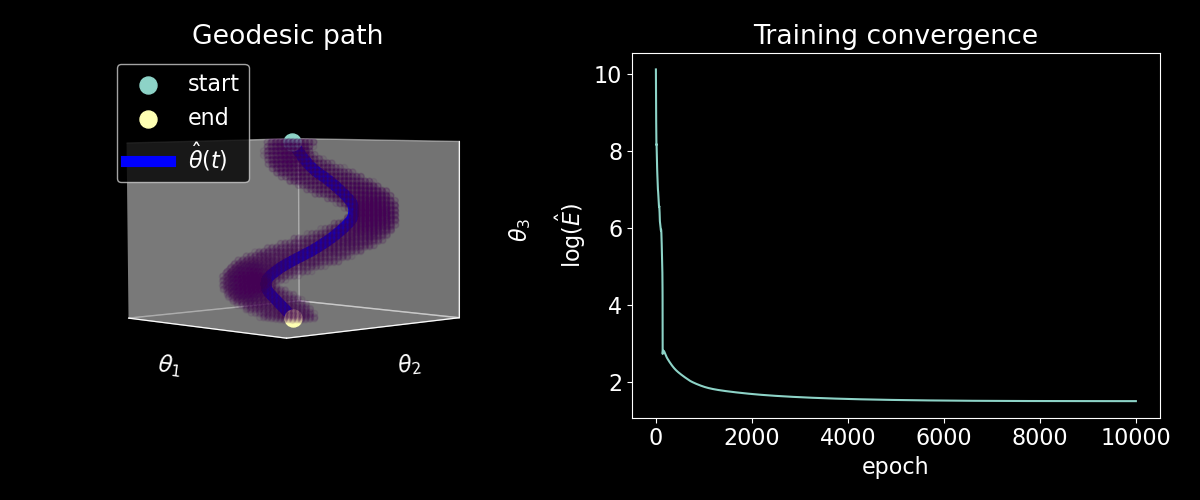}
\caption{When using embedded Fourier features, the optimizer manages to avoid the straight line local minimum, finding a winding path through the helical region of large wave speeds, which is shown in purple.}
\label{helix_path}
\end{figure}

\paragraph{} The particular Fourier feature embedding that we employ has $2f=30$ Fourier features and variance $\sigma^2=4$. The subsequent MLP network has two hidden layers with $15$ hidden units per layer. We take the refractive index of the helical tube to be $n_0=1$ (vacuum) and that of the surrounding medium to be $n_1=20$. See Figure \ref{helix_path} for the results. With the appropriate initialization, the parameterized path converges rapidly to wind through the helical tube.

\subsection{Minimal deformation of an elastic bar}

\paragraph{} For the third example, we take inspiration from solid mechanics to compute a minimal trajectory of a continuous function. Consider the displacement $u(x,t)$ of a linearly elastic bar defined on the unit interval $\Omega = [0,1]$ with unit material modulus $E=1$. The bar has homogeneous Dirichlet boundary conditions, that is, $u(0,t)=u(1,t)=0$. Our task is to find a trajectory $u(x,t)$ that takes the bar from an initial configuration $u_0(x) = u(x,0)$ to a final configuration $u_1(x)=u(x,1)$, minimizing an energy measure yet to be defined. In particular, we are interested in computing trajectories which smoothly interpolate between the two states. Thus, we wish to construct an energy functional to provide a measure of the total deformation in going between the two states. In solid mechanics, the elastic strain energy is often used as a measure of the severity of the deformation \cite{bendsoe_topology_2004}. The strain energy at time $t$ for the one-dimensional bar is given by 

\begin{equation*}
    U(t) = \int_0^1 \Psi\Big( u(x,t) \Big) dx = \int_0^1\frac{1}{2} \qty(\pd{u(x,t)}{x})^2 dx.
\end{equation*}

The rate of accumulation of strain energy is the power $P = \dot U$, which at first glance is a natural instantaneous measure of the deformation analogous to the velocity magnitude for a minimal distance geodesic. However, the variational objective arising from the power is

\begin{equation*}
    \int_0^1 P dt = \int_0^1 \dot U dt = U\Big( u(x,1) \Big) - U\Big( u(x,0) \Big).
\end{equation*}

Evidently, the power is a poor instantaneous measure of the deformation as it leads to an energy functional that depends only on the end states. We seek an energy functional whose minimum uniquely determines an interpolatory path between the initial and final states. One such energy functional is given by

\begin{equation}\label{deform_measure}
    E\Big( u(x,t) \Big)= \frac{1}{2} \int_0^1 \qty[\int_0^1 \qty(\frac{\partial^2 u(x,t)}{\partial x \partial t})^2dx ]dt.
\end{equation}

This energy functional penalizes changes in the strain at each point along the bar, thus encouraging smooth transitions between the two states. In order to discretize the spatial part of the problem, we represent the displacement field using an MLP with parameters $\boldsymbol \theta$. In particular, we have

\begin{equation}\label{net}
    \hat u(x;\boldsymbol \theta(t)) = \sin(\pi x)\mathcal{M}(x; \boldsymbol \theta(t) ),
\end{equation}

\noindent where the homogeneous Dirichlet boundary conditions on the displacement are enforced with a distance function. We remark that the network $\mathcal{M}$ is distinct from the discretization of the path through parameter space given in Eq. \eqref{distance}. Plugging the discretized displacement into Eq. \eqref{deform_measure}, the energy functional governing the minimal deformation trajectory becomes

\begin{equation*}
    E\Big( \boldsymbol \theta(t) \Big) = \frac{1}{2} \int_0^1 \dot \theta_k \dot \theta_j \qty[\int_0^1 \frac{\partial^2 \hat u}{\partial x \partial \theta_k} \frac{\partial^2 \hat u}{\partial x \partial \theta_j} dx]dt,
\end{equation*}

\noindent where the metric tensor is given by

\begin{equation}\label{strain_metric}
    g_{kj}(\boldsymbol \theta) = \int_0^1 \frac{\partial^2 \hat u}{\partial x \partial \theta_k} \frac{\partial^2 \hat u}{\partial x \partial \theta_j} dx.
\end{equation}

We now discretize the parameter trajectory $\boldsymbol \theta(t)$ with a second neural network. For clarity, we repeat the distance function parameterization given in Eq. \eqref{distance}:

\begin{equation*}
    \hat{\boldsymbol{\theta}}(t; \boldsymbol \beta) = \boldsymbol \theta_0(1-t) + \boldsymbol \theta_1 t + \sin(\pi t) \mathcal{N}(t;\boldsymbol \beta).
\end{equation*}

Note that the initial and final parameters $\boldsymbol \theta_0$ and $\boldsymbol \theta_1$ must be determined by training the network in Eq. \eqref{net} to match the initial and final displacement states with a standard regression problem. In other words, the parameters $\boldsymbol \theta_0$ and $\boldsymbol \theta_1$ are such that $\hat u(x; \boldsymbol \theta_0) = u_0(x)$ and $\hat u(x; \boldsymbol \theta_1) = u_1(x)$.

\paragraph{} Note that the metric tensor is a function of the displacement parameters $\boldsymbol \theta$, whose trajectory is discretized by another set of neural network parameters $\boldsymbol \beta$. As such, the metric tensor is $\mathbf{g} = \mathbf{g}(\boldsymbol \theta(\boldsymbol \beta))$. Using PyTorch, it is straightforward to use automatic differentiation to compute the entries of the metric tensor at given $\boldsymbol \theta$ with Eq. \eqref{strain_metric}. One must be careful, however, to ensure that the automatic differentiation of the objective in Eq. \eqref{drm} captures the dependence of $\boldsymbol \theta$ on $\boldsymbol \beta$ in the definition of the metric tensor. For example, writing a function that takes in parameters $\boldsymbol \theta$ to compute the metric will break the dependence $\boldsymbol \theta(\boldsymbol \beta)$ if the first step is to manually set the displacement network parameters to $\boldsymbol \theta$. The result will be gradient steps that do not correspond to descent directions of the objective.

\paragraph{} We discretize the displacement with a two hidden layer network $\mathcal{M}$ containing $10$ hidden units in each layer and hyperbolic tangent activation functions. This implies that the neural network approximation of the geodesic path in parameter space is a function $\mathbb{R}\rightarrow \mathbb{R}^{140}$, as $|\boldsymbol \theta |=140$ is the number of displacement parameters. The objective function is again that of Eq. \eqref{drm} with the metric given in Eq. \eqref{strain_metric}. We take the initial and final states as

\begin{equation*}
    u_0(x) = \sin(\pi x), \quad u_1(x) = - 2x\sin(\pi x).
\end{equation*}

\begin{figure}[hbt!]
\centering
\includegraphics[width=1.0\textwidth]{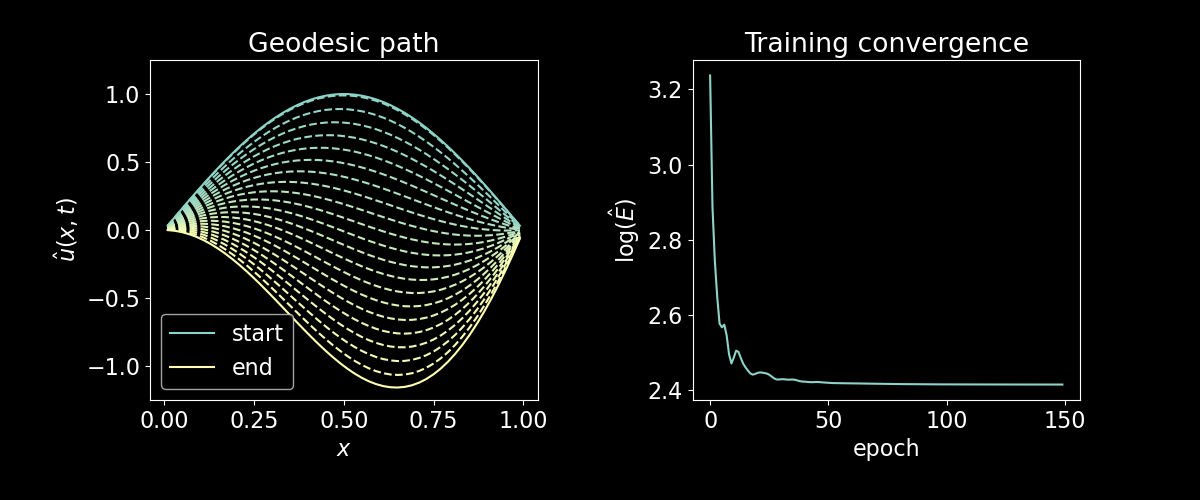}
\caption{Dashed lines indicate the trajectory of the displacement between the two boundary states (left). The objective function measuring the extent of deformation in transitioning between the two states rapidly converges to a minimum (right). }
\label{deform1}
\end{figure}

See Figure \ref{deform1} for the results. We remark that the intermediate states along the geodesic represent an appealing way to smoothly interpolate between the two functions. The converged value of the energy functional also acts as a distance measure between functions, as it measures the extent of deformation required to optimally transport one function to another. Given the connection to solid mechanics, geodesics of Eq. \eqref{strain_metric} are reminiscent of the continuum mechanics interpretation of the Wasserstein distance as the kinetic energy of the flow required to transport one distribution of mass to another \cite{brenier_computational_2000}.

\begin{figure}[hbt!]
\centering
\includegraphics[width=1.0\textwidth]{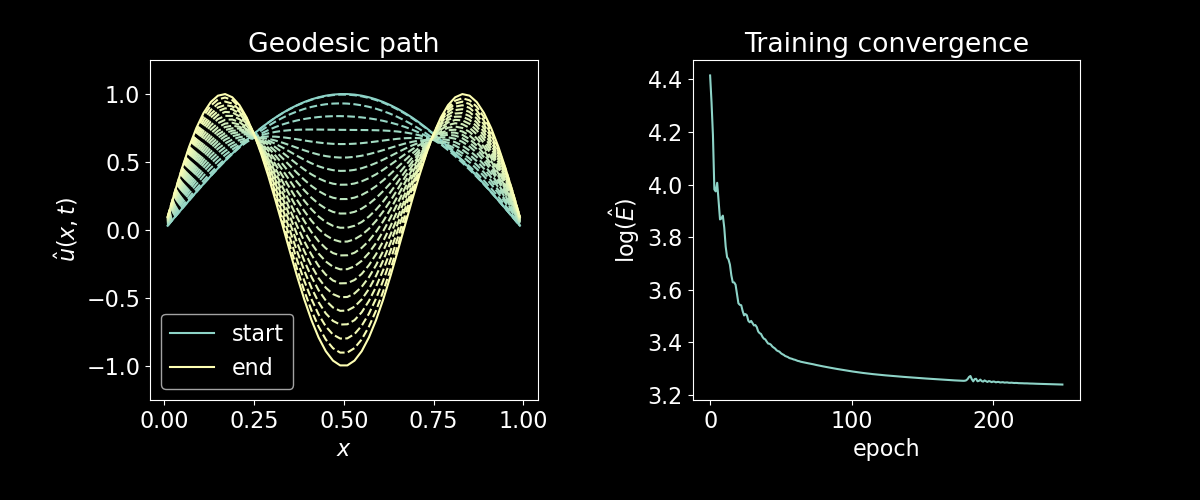}
\caption{Motion along the geodesic in the space of parameters discretizing the displacement represents a natural way to interpolate between states (left). The converged value of the energy functional also provides a measure of distance between two functions (right).}
\label{deform2}
\end{figure}

\paragraph{} We perform another test of the minimal deformation geodesic problem by changing the final state to $u_1(x) = \sin(3\pi x)$. See Figure \ref{deform2} for the optimal trajectory between these two states. It is interesting to note how the two points of intersection of the boundary states are stationary under the deformation. Once again, the path along the geodesic represents a technique for interpolating the two functions. We note that the converged value of the energy functional is larger in this second example, indicating that more deformation is required in transitioning between the two states.

\subsection{Geodesic interpolations in latent space}

\paragraph{} An interesting application of geodesics arises when interpolating in the latent space of generative models. Roughly speaking, a generative model learns a set of features that describe a data set and a decoder to perform reconstruction from the features. The features are assumed to follow a simple distribution, and thus can be sampled to generate data which is novel, yet statistically equivalent to the training data \cite{kingma_introduction_2019}. The low-dimensional space of features underlying the data is often referred to as the ``latent space" of the model. A common challenge with generative models is enforcing structure in the latent space, such that nearby points decode to similar quantities. Enforcing structure of this sort allows for the latent space to be interpolated, meaning that a path connecting two latent points corresponds to a smooth transition between the corresponding data. For example, interpolating the latent space of a generative model for images should result in smooth transitions from one image to another, with each intermediate image resembling data seen in training. This ability to interpolate is an intentional consequence of the convex support of the assumed distribution of the latent space. However, most interpolations take straight line paths between two latent points \cite{park_deepsdf_2019}. While structured latent spaces enforce similar decodings of closely spaced points, straight line interpolations do not necessarily correspond to ``minimal" transitions between the data at the two end points. Features may emerge then disappear in the intermediate decoded images, as seen in \cite{park_deepsdf_2019}. Analogous to the previous example, a geodesic path in the latent space can be found to minimize the ``deformation" in taking one image to another. The notion of deformation we adopt will be defined shortly.

\paragraph{} Recognizing that the latent space and decoder represent a parameterization of a data manifold, past works have found minimal interpolations using geodesics \cite{struski_feature-based_2024, stolberg-larsen_atlas_2022, hartwig_geodesic_2025}. In this example, we perform geodesic interpolation with a particular generative model and the DRM formulation of the geodesic problem, a combination which---to the best of our knowledge---has not been previously studied. We use a generative model designed for signed distance functions (SDFs) called ``DeepSDF" \cite{park_deepsdf_2019}. SDFs are scalar functions whose magnitude represents the shortest distance to the boundary of an object and whose sign indicates whether the spatial point is inside ($+$) or outside ($-$) the object. These functions are useful in rendering two- and three-dimensional geometries for applications such as computer graphics and topology optimization \cite{jiang_parametric_2017}. DeepSDF looks for a low-dimensional latent parameterization of the SDFs seen in training. The training data consists of a set of $N$ SDFs given by $\{ s_i(\mathbf{x}) \}_{i=1}^N$, where $\mathbf{x} \in \mathbb{R}^2$ in this example. Each SDF is given a latent code $\{ \mathbf{z}_i \}_{i=1}^N$ where $\mathbf{z}_i \in \mathbb{R}^{|z|}$. Next, an MLP decoder network $\mathcal{D}(\mathbf{z}, \mathbf{x} ; \boldsymbol \gamma)$ is introduced, with $\boldsymbol \gamma$ representing the trainable parameters of the network. The generative model is constructed by solving the following optimization problem: 

\begin{equation}\label{deepsdf}
\begin{aligned}
    \mathcal{L}( \boldsymbol \gamma , \{ \mathbf{z}_i \}_{i=1}^N) = \sum_{i=1}^N\qty( \lambda \lVert \mathbf{z}_i \rVert^2 + \int \Big( \mathcal{D}( \mathbf{x} ; \mathbf{z}_i ; \boldsymbol \gamma) - s_i(\mathbf{x})\Big)^2d\Omega) \\
    \boldsymbol \gamma^* , \{ \mathbf{z}^*_i \}_{i=1}^N = \underset{\boldsymbol \gamma , \{ \mathbf{z}_i \}_{i=1}^N}{\text{argmin }} \mathcal{L}.
\end{aligned}
\end{equation}

Note that the penalty on the latent code magnitudes enforces a normal prior on the latent distribution with variance determined by the penalty parameter $\lambda$. The second term in the loss enforces that the image of the latent variables under the decoder matches the training data. The primary insight of DeepSDF is to allow the latent variables themselves to be trainable parameters, thus avoiding the need for an encoding step like that of other popular generative models \cite{kingma_introduction_2019}. Once the neural network parameters $\boldsymbol \gamma$ are obtained, the latent space can be sampled to produce new SDFs. In practice, the integral in Eq. \eqref{deepsdf} is discretized on an integration grid, which we call $\mathcal{X} = \{ \mathbf{x}_j \}_{j=1}^Q$. Concerning ourselves only with the SDF on the integration grid, we denote the manifold defined by the trained decoder as

\begin{equation}\label{generative_manifold}
    \mathbf{N}( \mathbf{z}) = \mathcal{D}( \mathcal{X}, \mathbf{z}; \boldsymbol \gamma^*),
\end{equation}

\noindent where $\mathbf{N}: \mathbb{R}^{|z|} \rightarrow \mathbb{R}^Q$ maps from points in the latent space to the SDF evaluated on the integration grid. Parameterizing a path through the latent space of the decoder as $\mathbf{z}(t) = \boldsymbol \theta(t)$, finding a minimal interpolation of two SDFs with given latent codes is a straightforward application of the energy objective of Eq. \eqref{length}. This objective minimizes the accumulated change of SDF values at integration points in going from the initial to the final state. With the data manifold defined by Eq. \eqref{generative_manifold}, the metric tensor is given by Eq. \eqref{metric}, and the path in latent space is discretized per Eq. \eqref{distance}.

\paragraph{} To test DRM for geodesic interpolation in the latent space, we must first train the generative model. To do this, we generate SDFs corresponding to circles with varying positions and radii in the unit square $\Omega=[0,1]^2$. The SDF for a circle with radius $R$ centered at $\mathbf{x} =[a,b]^T$ is given analytically as 

\begin{equation*}
    s(\mathbf{x};R,a,b) = \sqrt{(x_1-a)^2 +(x_2-b)^2} - R.
\end{equation*}

\begin{figure}[hbt!]
\centering
\includegraphics[width=1.0\textwidth]{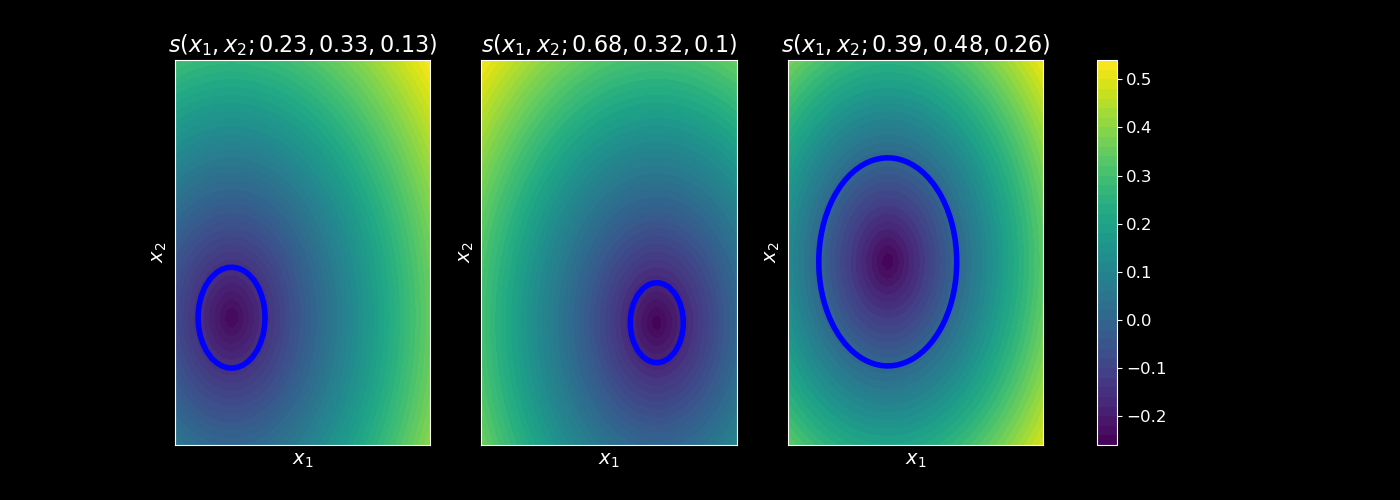}
\caption{The SDFs in the training data are generated by randomly sampling position and radii parameters. The radius is chosen such that the boundary of the circle lies within the unit square computational domain. The boundary is shown as the zero isocontour of the SDF in blue.}
\label{sdfs}
\end{figure}

With this in mind, we take the dimension of the latent space to be $|\mathbf{z}|=3$. For the training data, we generate $N=5000$ SDFs by independently randomly sampling the two coordinates $a$ and $b$ from a standard uniform distribution, then subsequently sampling the radius uniformly on an interval chosen such that the boundary of the circle is contained entirely within the unit square. See Figure \ref{sdfs} for examples SDFs from the training data. These fields are evaluated on a uniform integration grid of $Q=50 \times 50=2500$ points. The decoder MLP network uses hyperbolic tangent activation functions and consists of four hidden layers with a width of $20$ hidden units each. The penalty parameter is taken to be $\lambda= 5 \times 10^{-4}$, which is observed empirically to encourage latent distributions of approximately unit variance. A stochastic version of Eq. \eqref{deepsdf} is employed in practice, in which a batch of $250$ SDFs and $500$ integration points are used to compute the objective and its gradient at each optimization step. We use ADAM optimization with a learning rate of $7.5 \times 10^{-3}$ and run the optimization for $3\times 10^5$ steps.

\paragraph{} Once the model is trained, we can test its behavior when interpolating by generating SDFs with Eq. \eqref{generative_manifold}. The path we take in the latent space is given by 

\begin{equation*}
    \boldsymbol \theta(t) = 0.7\begin{bmatrix}
        -1 \\ -1 \\ 1
    \end{bmatrix}(1-t) + 0.5\begin{bmatrix}
        1 \\ 1 \\ -1
    \end{bmatrix} t.
\end{equation*}

\begin{figure}[hbt!]
\centering
\includegraphics[width=1.0\textwidth]{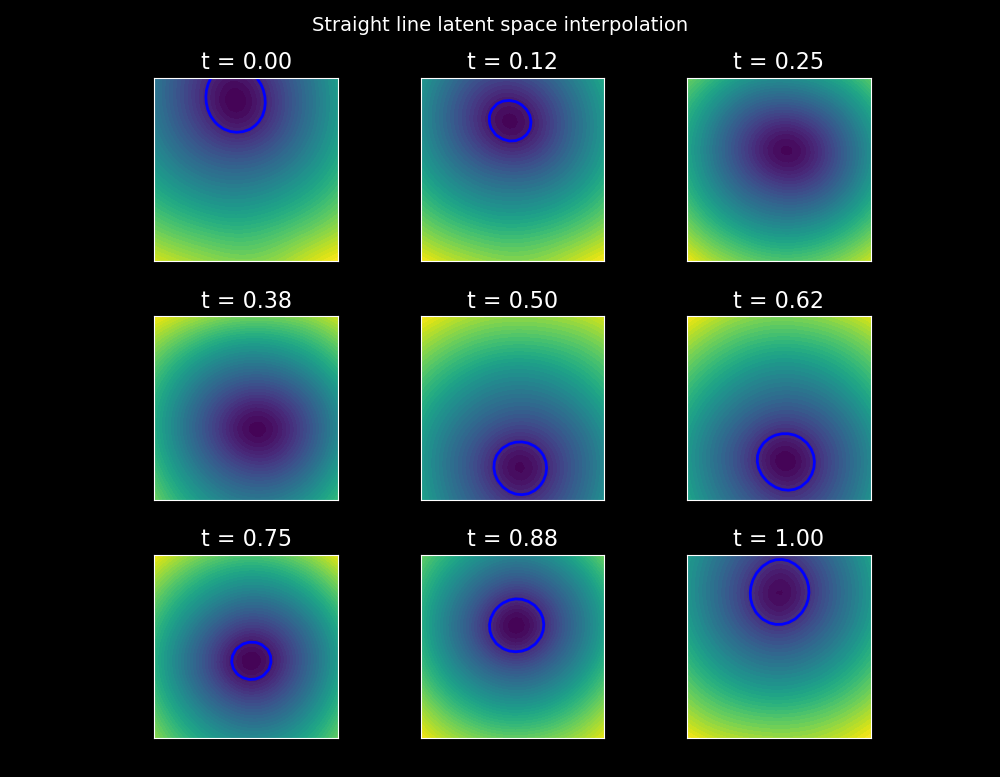}
\caption{Straight line interpolation in the latent space does not correspond to a minimal trajectory in physical space. The centers of the intermediate circles drift away from the final circle, only to return later. Furthermore, the circle is closed and then re-opened. A geodesic in latent space is intended to mitigate this kind of superfluous motion.}
\label{interp}
\end{figure}

\begin{figure}[hbt!]
\centering
\includegraphics[width=1.0\textwidth]{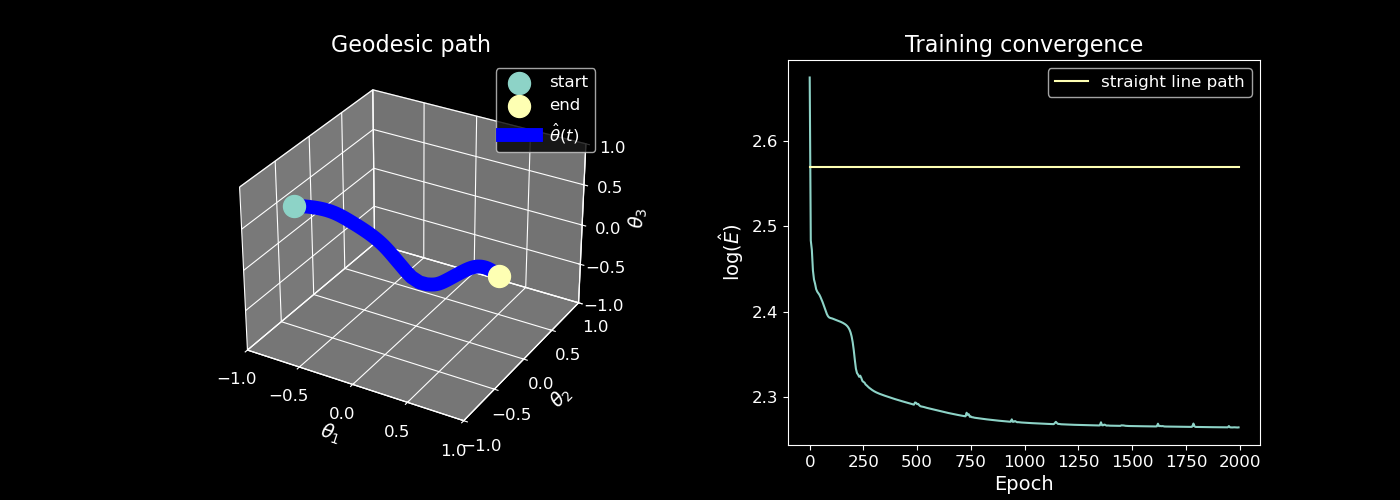}
\caption{The minimal interpolation of the two SDFs does not follow a straight line (left). The decrease in the objective is approximately monotonic, but saturates after $2000$ steps. The energy of the geodesic path is approximately $70\%$ that of the straight line interpolation.}
\label{latent_path}
\end{figure}

\begin{figure}[hbt!]
\centering
\includegraphics[width=1.0\textwidth]{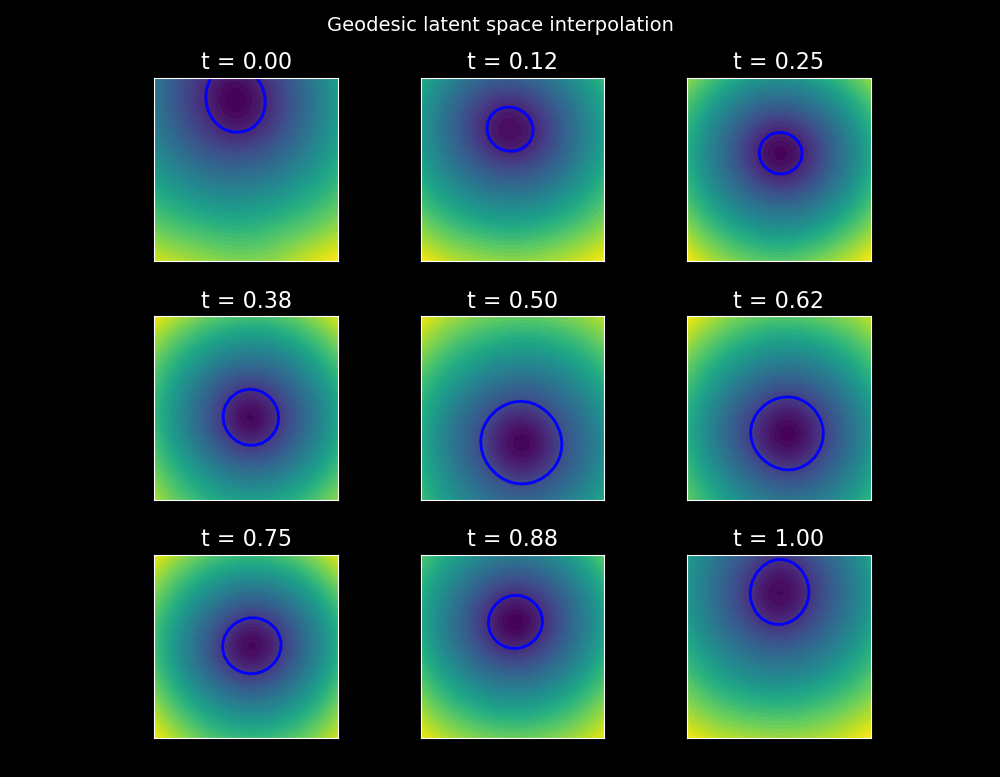}
\caption{The geodesic interpolation mitigates the drift of the circle's center from the initial and final states, as well as entirely preventing the zero radius intermediate states obtained from the straight line interpolation.}
\label{interp_geodesic}
\end{figure}

See Figure \ref{interp} to visualize the decoded SDFs at discrete points along the straight line path in the latent space. These snapshots show that the position and radii of the circle do not monotonically approach their final values. Intuitively, a minimal path between the initial and final states would correspond to straight line motion of the center of the circle and a monotonic increase or decrease of the radius. The solution to the geodesic problem will reveal if such a path is possible with the given latent space structure. To this end, we discretize the path with a two hidden layer network of width $25$. The metric tensor is computed using automatic differentiation of the decoder neural network with respect to the input latent coordinates. We run the optimization of Eq. \eqref{drm} for $2000$ epochs. See Figure \ref{latent_path} for the geodesic path and the training convergence. The geodesic is seen to deviate from the straight line path. The modest decrease in the objective indicates that it is not possible to entirely avoid the superfluous motion of the SDF. Figure \ref{interp_geodesic} shows snapshots of the SDFs along the geodesic trajectory. Though the center of the circle still performs a loop, it remains closer to the initial and final centers than the straight line interpolation. Additionally, in minimizing the accumulated change to the SDF, the geodesic path never closes the circle. This is a significant qualitative difference from the straight line interpolation. 

\section{Conclusion}

\paragraph{} From path planning to optics, and solid mechanics to generative modeling, geodesic boundary value problems are ubiquitous in physics and engineering. Despite the recent explosion of interest in neural network discretizations of differential equations, geodesic problems have yet to be investigated in the SciML literature. Due to their time-like structure, these problems have simple geometry. Though existing approaches often solve the Euler-Lagrange equations corresponding to stationarity of the energy functional, geodesic problems are variational. Furthermore, the variational optimization problem is naturally nonlinear. Any problem that satisfies these three criteria is well-suited to the Deep Ritz method, and indeed, we have shown that applying this tool set to geodesic problems yields an efficient and straightforward solution strategy. One noteworthy advantage of using the Deep Ritz method over a shooting method is that the initial and final conditions can be satisfied by construction, so no iteration of initial conditions is required.

\paragraph{} This work represents only a simple proof of concept for solving geodesic problems with the Deep Ritz method. While PINN-based approaches have been successful on a wide array of problems, they often introduce significant computational cost which may not be justified by improvements in solution accuracy. This is especially the case for linear problems---such as those of elasticity and heat transfer---in which accurate solutions can be rapidly and reliably obtained on complex three-dimensional geometries with legacy tools such as the finite element method. It is difficult to imagine how a machine learning-based approach could outperform solution strategies of this sort. There are, however, some problems which, by all appearances, are fine-tuned for machine learning methods. The three characteristics we have identified above represent our attempt to argue that geodesic problems are one such example. Future work will focus on Deep Ritz solutions to geodesic problems of increased scale and complexity. The author suspects that a vanilla Deep Ritz implementation may, in some circumstances, outperform standard shooting methods in both computational cost and accuracy. Confirmation or denial of such unabashed speculation must await further studies.



\end{document}